\documentclass[10pt,twocolumn,letterpaper]{article}

\usepackage[pagenumbers]{cvpr} 
\usepackage{multirow}
\usepackage{makecell}
\usepackage{adjustbox}
\usepackage{array}
\usepackage{xcolor,colortbl}
\usepackage{algorithm}
\usepackage{algpseudocode}
\usepackage{multicol}
\usepackage{caption}
\usepackage{graphicx}
\usepackage{subcaption}
\usepackage{wrapfig}
\usepackage{amsmath}
\usepackage{bbm}

\newcolumntype{R}[2]{%
    >{\adjustbox{angle=#1,lap=\width-(#2)}\bgroup}%
    l%
    <{\egroup}%
}

\definecolor{lightgrey}{rgb}{0.95, 0.95, 0.95}
\definecolor{softblue}{rgb}{0.22, 0.59, 1.00}
\definecolor{softorange}{rgb}{1.00, 0.52, 0.00}
\definecolor{softgreen}{rgb}{0.45, 0.62, 0.5}
\definecolor{lightorange}{RGB}{255, 223, 186}

\usepackage{multirow}
\usepackage{colortbl}
\usepackage[dvipsnames]{xcolor}
\usepackage{makecell}
\usepackage{rotating}

\usepackage{amsmath,amsfonts,bm}

\def\eqref#1{equation~\ref{#1}}

\def\1{\bm{1}}

\DeclareMathAlphabet{\mathsfit}{\encodingdefault}{\sfdefault}{m}{sl}
\SetMathAlphabet{\mathsfit}{bold}{\encodingdefault}{\sfdefault}{bx}{n}

\newcommand{\bh}{\mathbf h}

\newcommand{\bw}{\mathbf w}

\newcommand{\bs}{\mathbf s}

\newcommand{\bX}{\mathbf X}

\newcommand{\bW}{\mathbf W}
\newcommand{\bA}{\mathbf A}

\newcommand{\bB}{\mathbf B}

\newcommand{\mcD}{\mathcal D}

\newcommand{\mcS}{\mathcal S}

\newcommand{\mcE}{\mathcal E}

\newcommand{\mbbR}{\mathbb R}

\definecolor{cvprblue}{rgb}{0.21,0.49,0.74}
\usepackage[pagebackref,breaklinks,colorlinks,allcolors=cvprblue]{hyperref}

\usepackage{xcolor}

\def\eg{\textit{e.g.}}
\def\ie{\textit{i.e.}}

\title{On Token's Dilemma: Dynamic MoE with Drift-Aware Token Assignment for Continual Learning of Large Vision Language Models}

\author{
Chongyang Zhao\quad
Mingsong Li\quad
Haodong Lu\quad
Dong Gong\thanks{D. Gong is the corresponding author.}\\
{University of New South Wales (UNSW Sydney)}\\
\tt{\small \{chongyang.zhao, dong.gong\}@unsw.edu.au}\\
}

\begin{document}

\maketitle
\begin{abstract}
Multimodal Continual Instruction Tuning aims to continually enhance Large Vision Language Models (LVLMs) by learning from new data without forgetting previously acquired knowledge. Mixture of Experts (MoE) architectures naturally facilitate this by incrementally adding new experts and expanding routers while keeping the existing ones frozen. However, despite expert isolation, MoE-based continual learners still suffer from forgetting due to routing-drift: old-task tokens become mistakenly attracted to newly added experts, degrading performance on prior tasks. We analyze the failure mode at the token level and reveal the token's dilemma: ambiguous and old tokens in new-task data offer minimal learning benefit yet induce forgetting when routed to new experts, due to their ambiguous routing assignment during training. Motivated by this, we propose LLaVA-DyMoE, a dynamic MoE framework that incrementally expands the MoE with drift-aware token assignment. We characterize token types via their routing score distributions and apply targeted regularization. Specifically, a token-level assignment guidance steers ambiguous and old tokens away from new experts to preserve established routing patterns and alleviate routing-drift, while complementary routing score regularizations enforce expert-group separation and promote new-expert specialization. Extensive experiments demonstrate that our LLaVA-DyMoE effectively mitigates routing-drift-induced forgetting, achieving over a 7\% gain in mean final accuracy and a 12\% reduction in forgetting compared to baselines. The project page is \href{https://zhaoc5.github.io/DyMoE}{\texttt{zhaoc5.github.io/DyMoE}}.
\end{abstract}

\section{Introduction}\label{sec:intro}
Large Vision Language Models (LVLMs)~\cite{liu2023visual,wang2024emu3,bai2025qwen2,comanici2025gemini} have recently achieved remarkable performance across a wide range of vision-language tasks~\cite{lin2014microsoft,antol2015vqa,mishra2019ocr} by extending Large Language Models (LLMs)~\cite{touvron2023llama,jiang2023mistral7b,qwen2.5} to process multimodal information.
Central to their success is a two-phase development pipeline: pre-training for vision-language alignment, followed by instruction tuning to adapt the model to specific domains and tasks. 
While these models are trained on fixed datasets and remain largely static, new instruction-following requirements often arise dynamically in real-world applications.
This motivates continual learning capabilities that allow the model to assimilate new knowledge while preserving performance on previously learned tasks, overcoming catastrophic forgetting~\cite{french1999catastrophic,li2017learning,kirkpatrick2017overcoming}.

\begin{figure}[!t]
    \centering
    \includegraphics[width=0.8\linewidth]{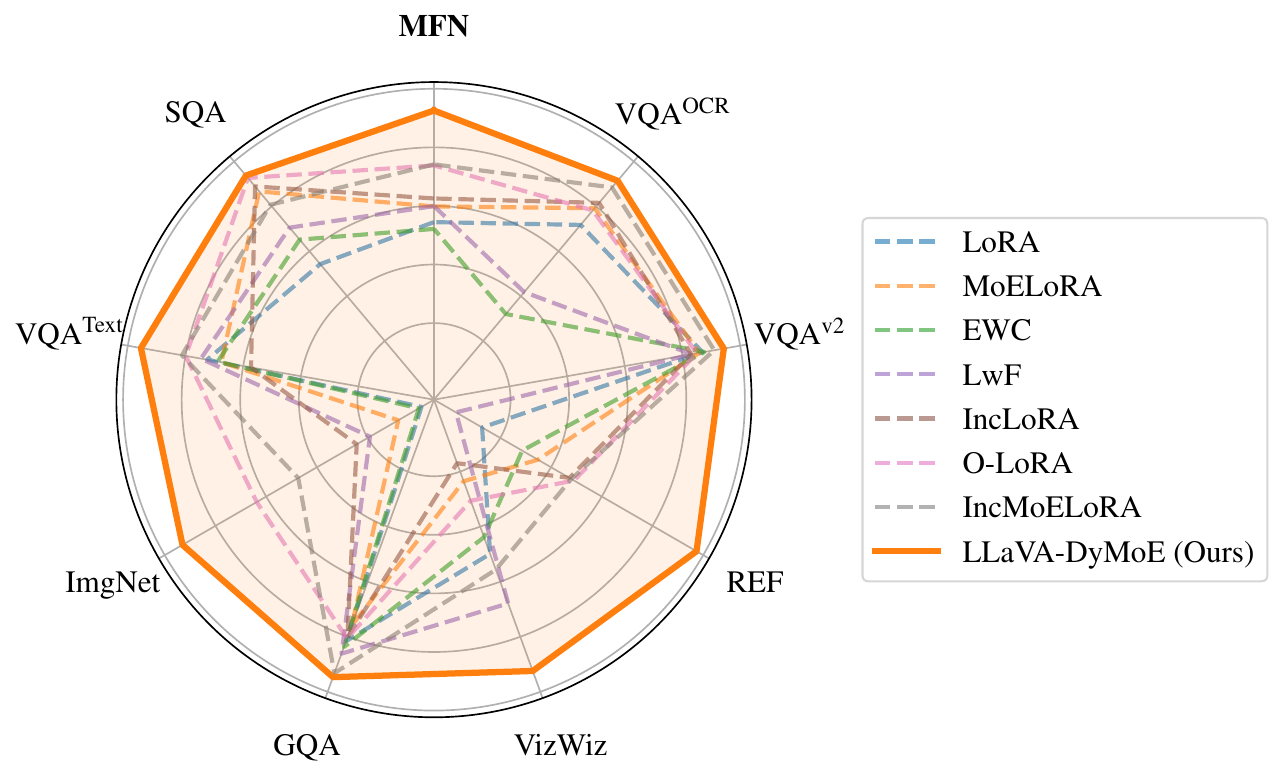}
\caption{Performance comparison on the CoIN benchmark, showing per-task final accuracy and mean final accuracy (MFN).}
\label{fig:radar}
\vspace{-0.6cm}
\end{figure}

As naively retraining on a combined set of old and new data is resource-intensive, Multimodal Continual Instruction Tuning (MCIT)~\cite{he2023continual, cao2024continual, chen2024coin, chen2025sefe, yu2025progressive} has emerged to address this need, aiming to incrementally instruction-tune LVLMs on new tasks while maintaining proficiency on previously learned ones. 
Common strategies include regularization-based methods~\cite{he2023continual,lao2023multi,chen2025sefe,qiao2024large}, which impose parameter constraints to prevent forgetting, and rehearsal-based methods~\cite{lei2023symbolic,marouf2025ask,yu2025progressive}, which rely on replaying old data. However, these approaches often introduce non-trivial computational overhead or storage constraints.
An attractive alternative is to isolate task-specific parameters via parameter-efficient tuning (PEFT) approaches~\cite{hu2022lora, fedus2022switch, zeng2024modalprompt,chen2025sefe}. Among these, the Mixture of Experts (MoE) paradigm~\cite{fedus2022switch,zadouri2024pushing,yang2025solving,yu2025progressive,wang2025smolora} has become a prevalent solution owing to its dynamic modular architecture, superior scalability, and inference efficiency. This modular structure inherently facilitates flexible expert allocation and parameter isolation across tasks, which is crucial for mitigating catastrophic forgetting. 

Despite these advantages, existing MoE-based MCIT approaches still exhibit significant forgetting. Training a fixed-size MoE without parameter isolation across shared experts and routers leads to inter-task interference and degraded knowledge retention~\cite{chen2024coin}.
Some works~\cite{zhao2025llava,yu2025progressive,guo2025hide,huai2025cl} address this by incrementally expanding model experts while freezing previous ones, and introducing task-specific routers to identify tasks and assign experts accordingly.
However, reliable task identification may require heavy computation and can become unreliable when tasks are diverse and complex.
Moreover, task-level expert assignment reduces the combinatorial flexibility of MoE, sacrificing its inherent token-level routing adaptability.

\begin{figure*}[t]
    \centering
    \subfloat[Only retain new tokens]{
    \includegraphics[width=0.27\linewidth]{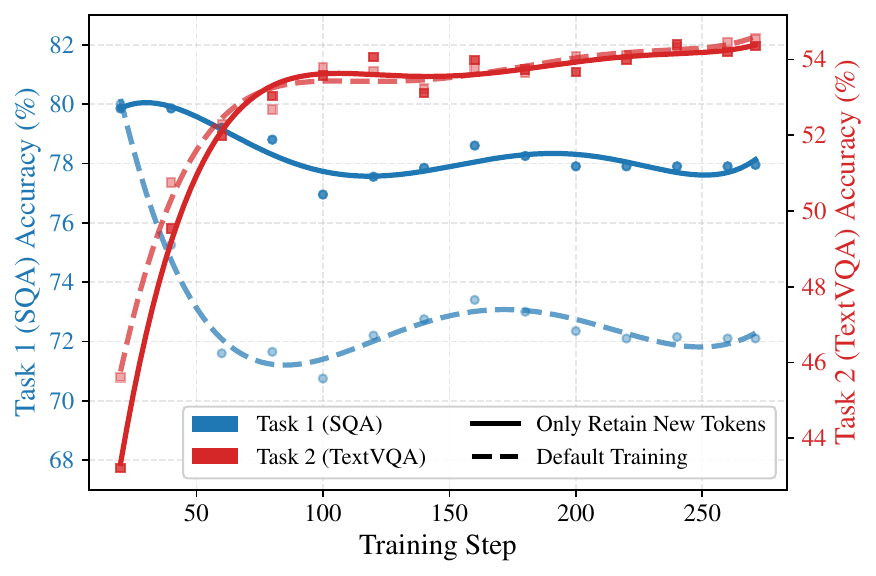}
        \label{fig:exp_new}
    }
    \hfill
     \subfloat[Mask out old tokens]{
        \includegraphics[width=0.27\linewidth]{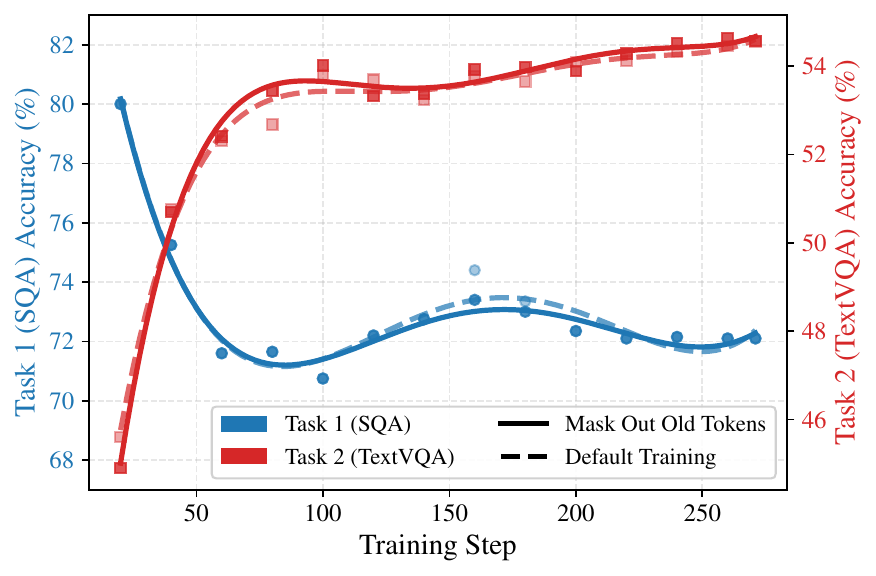}%
        \label{fig:exp_no_old}
    }
    \hfill
    \subfloat[Only retain ambiguous tokens]{
        \includegraphics[width=0.27\linewidth]{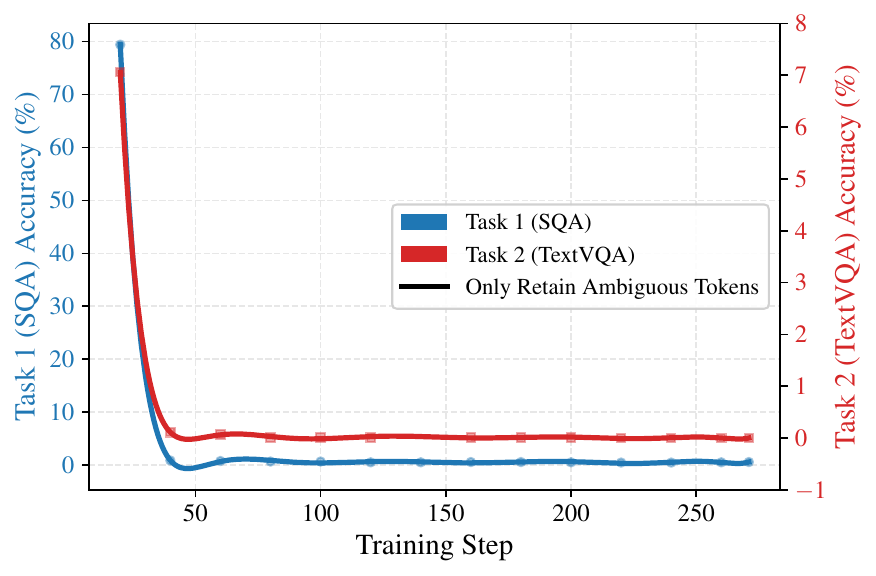}
        \label{fig:exp_ambi}
    }
    \vspace{-0.1cm}
\caption{Routing-drift analysis in a controlled two-task learning experiment. After learning on the 1st task (SQA), we conduct the 2nd task (TextVQA) training using the baseline (default training) and three different token masking strategies based on token type. Throughout the training stages, we evaluate forgetting (decrease in task 1 accuracy) and new-task learning (improvement of task 2 accuracy). Polynomial regression-fitted curves are used for better visualization and readability of performance changes. The baseline (default training) refers to a scenario where each input token is assigned to all experts (including old frozen and new learnable ones). We then examine the role of each token group based on its routing score: (a) We only retain the contribution of tokens that have high affinity to the new expert group (termed ``new tokens''). (b) We mask out the contribution of tokens with a high affinity to the old expert group (termed ``old tokens''). (c) We only retain the contribution of tokens with a small affinity difference between the old and new expert group (termed ``ambiguous tokens'').}
    \label{fig:analyses}
    \vspace{-0.2cm}
\end{figure*}

In this paper, we introduce LLaVA-DyMoE, a Dynamic MoE framework with Drift-Aware Token Assignment for continual learning of LVLMs.
Unlike prior works that bypass routing instability via task-specific routing~\cite{yu2025progressive,guo2025hide,huai2025cl}, we focus on directly addressing the underlying token-level cause of forgetting during dynamic MoE expansion.
Even with old experts and their router parameters frozen, training the newly added components on new-task data still causes forgetting: the updated routing parameters distort the assignment of old-task tokens to their established experts.
This distortion constitutes \textbf{routing-drift}, a corruption of the router's learned policy for old tasks that drives catastrophic forgetting at the token level. 
We analyze how routing-drift arises during training of newly added components and reveal that not all tokens in the new-task data contribute equally (Sec.~\ref{sec:analyze} and Fig.~\ref{fig:analyses}).
Beyond \textit{new tokens} that carry genuinely novel patterns, we identify two types that pose a forgetting risk: \textbf{\textit{ambiguous tokens}}, which exhibit similar routing affinity for both old and new expert groups; and \textbf{\textit{old tokens}}, whose patterns closely resemble old tasks yet receive non-negligible new-expert weight from the under-optimized router.
Both types offer minimal benefit for new-task learning, yet when routed to new experts, they inadvertently train the new router to attract old-task patterns---causing old-task tokens to be mis-routed at inference time and inducing forgetting. This is the \textit{token's dilemma}: minimal learning value, yet a direct forgetting cost when left unguided; compounded by the inherent ambiguity of their routing and expert assignment.

Motivated by this analysis, LLaVA-DyMoE mitigates forgetting in dynamic MoE expansion through a two-fold regularization comprising Token Assignment Guidance (TAG) and Routing Score Regularization (RSR).
TAG identifies token types from their routing scores and guides their assignment by adjusting routing scores during training, directly tackling the tokens' dilemma and steering ambiguous tokens away from new experts.
As a complementary soft regularization, RSR encourages exclusive token-to-group routing and promotes new-expert specialization on genuinely new-task tokens.
Extensive experiments on the CoIN benchmark~\cite{chen2024coin} across eight VQA-based tasks demonstrate the effectiveness of LLaVA-DyMoE, achieving over a 7\% gain in MFN and a 12\% reduction in forgetting compared to baseline methods.

Moreover, LLaVA-DyMoE is orthogonal to and compatible with existing MCIT paradigms, including data-based methods~\cite{rolnick2019experience,yu2025progressive,chen2025sefe} and task-specific routing approaches~\cite{yu2025progressive,guo2025hide,yu2024boosting,zhao2025llava}, and can be combined with them for further performance gains.

Our main contributions are summarized as follows:
\begin{itemize}
\item We identify the token-level cause of \textit{routing-drift}: the \textit{token's dilemma}. Through controlled analysis, we show that \textit{ambiguous tokens} and \textit{old tokens} in new-task data offer minimal new-task benefit yet induce forgetting when routed to new experts. Ambiguous tokens are especially challenging, as their ambiguous affinity makes them difficult to identify and prone to unstable routing. (Sec.~\ref{sec:analyze})

\item Motivated by this, we introduce LLaVA-DyMoE, a two-fold regularization framework. It comprises a Token Assignment Guidance (TAG) mechanism that identifies and redirects ambiguous tokens away from new experts, and a Routing Score Regularization (RSR) that encourages exclusive token-to-group routing and promotes new-expert specialization. (Sec.~\ref{sec:meth-DATAR})

\item Extensive experiments demonstrate that our method significantly outperforms baseline methods, achieving a superior balance between knowledge retention and new-task acquisition. (Sec.~\ref{sec:exper})
\end{itemize}

\section{Related Work}
\textbf{Continual Learning} (CL) investigates methods for training models on non-stationary data distributions, typically presented as a sequence of tasks, with the primary goal of overcoming catastrophic forgetting~\cite{de2021continual,wang2024comprehensive,zhao2024learning,rebuffi2017icarl,kirkpatrick2017overcoming,yoon2017lifelong}. 
CL methods can be broadly categorized by their core strategy to mitigate catastrophic forgetting. 
Rehearsal-based methods~\cite{rebuffi2017icarl,lopez2017gradient,chaudhry2019tiny,buzzega2020dark,shin2017continual,rostami2019complementary,riemer2019scalable} store and generate a small subset of previous samples or features during training on new tasks, thereby approximating the data distribution of the past.
Regularization-based methods~\cite{kirkpatrick2017overcoming,zenke2017continual,nguyen2017variational,li2017learning,aljundi2018memory,zhang2020class,zhang2023slca,jha2024clap4clip} mitigate catastrophic forgetting by penalizing updates to parameters deemed critical for performance on previous tasks.
Architecture-based methods~\cite{yoon2017lifelong,serra2018overcoming,li2019learn,yan2021dynamically,ye2023self,yu2024boosting,wang2025self,lu2024adaptive,lu2025little} allocate new parameters for each task, either by physically expanding the network or by functionally isolating parameter subsets via masking.

\noindent\textbf{Continual Learning for LVLMs and LLMs.}
Continually expanding the capabilities of LLMs~\cite{touvron2023llama,jiang2023mistral7b,qwen2.5} and LVLMs~\cite{liu2023visual,wang2024emu3,bai2025qwen2,comanici2025gemini} presents unique challenges, as the immense computational cost of retraining makes continual instruction tuning a necessity. In the vision-language domain, recent efforts~\cite{he2023continual,chen2024coin,qiao2024large,guo2025hide,wang2025smolora,huai2025cl,zhang2025enhancing,ge2025dynamic,cao2024continual} focus on continual instruction-tuning LVLMs with sequential tasks, avoiding the expensive process of retraining from scratch. MoELoRA~\cite{chen2024coin} proposes the CoIN benchmark and adopts the framework of Mixture of Experts (MoE) with LoRA experts. SEFE~\cite{chen2025sefe} incrementally learns new LoRA matrices and regularizes key parameter updates to retain prior knowledge. ProgLoRA~\cite{yu2025progressive} proposes a progressive LoRA pool that mitigates task interference by isolating knowledge in separate LoRA blocks. In the language domain, similar efforts have been applied to either regularize learning~\cite{wang2023orthogonal,razdaibiedina2023progressive} or expand the capacity of the model~\cite{razdaibiedina2023progressive}.

\noindent\textbf{Mixture of Experts (MoE) with LoRA.}
The MoE paradigm enhances model capacity by replacing the Transformer's dense feed-forward layer with multiple expert subnetworks and a routing network~\cite{shazeer2017outrageously,lepikhin2020gshard,fedus2022switch,dai2024deepseekmoe,yang2025solving,guo2024dynamic}. This framework dynamically routes each input to a sparse subset of experts, employing auxiliary load-balancing losses~\cite{fedus2022switch} to ensure balanced expert utilization. This paradigm has been adopted in conjunction with LoRA~\cite{hu2022lora} for standard fine-tuning~\cite{chen2023octavius,li2024mixlora,dou2023loramoe} and for continual learning~\cite{yu2024boosting,chen2024coin}, where low-rank adapters are treated as experts. Our formulation adopts this MoE with LoRA paradigm, where we add new LoRA experts for each task to expand the knowledge base of the foundation model.

\section{Methodology: LLaVA-DyMoE with Drift-Aware Token Assignment}\label{sec:method-llava-dymoe}
\subsection{Problem Setup}
Continual Learning aims to enable models to continually acquire new knowledge without catastrophic forgetting.
Within the broader CL paradigm, Multimodal Continual Instruction Tuning (MCIT) enables Large Vision-Language Models (LVLMs) to incrementally adapt to new tasks and maintain strong performance on previously learned tasks, without full retraining.
Let $\{\mcD_1, ..., \mcD_t, ..., \mcD_T \}$ denote the training data of a sequence of $T$ tasks arriving as a stream.
The dataset $\mcD_t=\{\bX\}^{S_t}$ for the $t$-th task consists of $S_t$ samples. Each sample is a multimodal instruction-response triplet $\mathbf{X} = (\mathbf{x}_{v}, \mathbf{x}_{q}, \mathbf{x}_{a})$. Here, $\mathbf{x}_{v}$, $\mathbf{x}_{q}$, and $\mathbf{x}_{a}$ denote the image, instruction, and answer tokens, respectively. 
We focus on the MCIT setting~\cite{chen2024coin} based on LLaVA~\cite{liu2023visual}.

\subsection{Dynamic MoE Layers}
\label{sec:dymoe_layer}
\noindent \textbf{MoE Layers with LoRA.} 
Given a pre-trained LLaVA, learning on new instruction tuning tasks is achieved through fine-tuning with LoRA \cite{hu2022lora}. 
A LoRA module parameterizes a low-rank update to a pre-trained weight matrix $\mathbf W_0^{l,m} \in\mathbb R^{d_\text{out} \times d_\text{in}}$ (at layer $l$, module $m$ in the Transformer of LLaVA) by introducing two factors $\mathbf{B}^{l,m} \in \mathbb{R}^{d_\text{out} \times r}$ and $\mathbf{A}^{l,m} \in \mathbb{R}^{r \times d_\text{in}}$, such that $\Delta \bW^{l,m} = \bB^{l,m} \bA^{l,m}$, where $r \ll \min(d_\text{in},d_\text{out})$. The updated weight matrix is then defined as: $\mathbf{W}^{l,m} = \mathbf{W}_0^{l,m} + \Delta \mathbf{W}^{l,m} = \mathbf{W}_0^{l,m} + \mathbf{B}^{l,m} \mathbf{A}^{l,m}.$

Instead of relying on a single continually trained LoRA adapter or merging task-specific adapters into the backbone,
we develop an MoE architecture with LoRA modules as experts to augment each module with weight matrix $\bW_0^{l,m}$ in the pre-trained LLM of LLaVA. 
An MoE layer is composed of multiple experts $\{e^{l,m}_i()\}^N_{i=1}$ and a router $R^{l,m}():\mbbR^{d_\text{in}}\rightarrow \mbbR^N$ to assign each input token representation to specific experts. Each expert $e_i()$ is a LoRA module parameterized by $(\mathbf{A}_i^{l,m}, \mathbf{B}_i^{l,m})$. 
Given an input multimodal token's representation $\mathbf{h}^{l,m} \in \mathbb{R}^{d_\text{in}}$, the output $\bh^{l,m}_\text{out}$ is computed as:
\begin{equation}\label{eq:moelora}
\begin{split}
& \mathbf{h}^{l,m}_{\text{out}} = \mathbf{W}_0^{l,m}\mathbf{h}^{l,m} + \sum\nolimits_{i=1}^{N} {w}_i \mathbf{B}_i^{l,m}\mathbf{A}_i^{l,m}\bh^{l,m},\\
& w_i = w'_i/{\sum\nolimits_{i=1}^N w'_i},~w'_i = \exp(s_i)\mathbbm{1}[i\in \mathtt{TopK}_K(\bs)],
\end{split}
\end{equation}
where $\mathbf{B}_i^{l,m}\mathbf{A}_i^{l,m}\bh^{l,m}=e_i(\bh^{l,m})$, $\bs=R^{l,m}(\bh^{l,m})$ is the logits vector produced by the router, $\mathtt{TopK}_K(\bs)$ denotes the set comprising the indices of the $K$ highest affinity scores among all $N$ experts, $\mathbbm{1}()$ is the indicator function, $s_i$ and $w_i$ are the $i$-th elements of $\bs$ and $\bw$, respectively. 
The router assigns the token to the corresponding $K$ experts with top-$K$ highest scores. 
The resulting routing weight $\bw$ is sparse, indicating that only $K$ out of $N$ gate values are nonzero. This sparsity property encourages the tokens to be assigned to specialized experts at each layer. 

\noindent \textbf{Dynamic MoE with incrementally added experts.}
We implement dynamic MoE (DyMoE) with LoRA experts, incrementally adding new experts and expanding the router as each new task arrives.
In MCIT, when the $t$-th task ($t>1$) arrives, we assume existing experts $\mcE_{t-1}$ with a router producing scores $\bw_{t-1}, \bs_{t-1}\in\mbbR^{|\mcE_{t-1}|}$, indexed by $\mcS_{t-1}$.
For a new task $t$, we add $N_t$ new experts $\mcE_{t, \text{new}}=\{e^{l,m}_i\}^{N_t}_{i=1}$ and expand the router to produce $N_t$ new output routing scores $\bw_{t,\text{new}}, \bs_{t,\text{new}}\in\mbbR^{N_t}$.
All old existing parameters are frozen; only the newly added experts $\mcE_{t,\text{new}}$ and their associated router parameters are trained, while input tokens from task $t$ can be routed to both old frozen experts $\mcE_{t-1}$ and new trainable experts $\mcE_{t,\text{new}}$.
After adding and training new experts, the expert set and router output are expanded: $\mcE_{t}=\mcE_{t-1}\cup \mcE_{t, \text{new}}$, $\bw_t=[\bw_{t-1}; \bw_{t, \text{new}}]$, and $\bs_t=[\bs_{t-1}; \bs_{t, \text{new}}]$, and the index set is updated as $\mcS_{t}=\mcS_{t-1}\cup \mcS_{t, \text{new}}$. 

\noindent \textbf{Forgetting as routing-drift in DyMoE.}
During training, newly added experts and their router parameters are updated while existing experts remain frozen, allowing new-task tokens to reuse previously learned knowledge and keeping experts isolated across tasks.
However, despite this isolation, DyMoE still exhibits catastrophic forgetting in MCIT due to \emph{routing-drift}. After updating the new router parameters, old-task tokens may be mis-routed to newly added experts that were never trained on them, resulting in performance degradation on old tasks, \ie, \emph{forgetting}. 

\subsection{Token’s Dilemma: Analyses on Routing-drift Associated with Token Assignment}
\label{sec:analyze}
Although many CL and MCIT methods with incrementally added network components \cite{zhao2025llava,yu2025progressive,wang2025self,zhang2025enhancing,chen2023lifelong} attempt to handle or bypass the forgetting caused by routing-drift, 
they typically rely on auxiliary mechanisms, such as task-specific router predictors or auxiliary regularizers. 
Instead, we aim to investigate and tackle the inherent cause of routing-drift in the dynamic MoE expansion process. We analyze how routing-drift is caused at the token level when only newly added parameters are updated on new-task data while old parameters remain frozen. 
Even when trained only on new-task tokens, the newly added router parameters can still attract old-task tokens (\ie, high routing scores $\bs_{t, \text{new}}$) and route them to new experts. 
Since MoE routers operate on and are also trained by individual token-expert assignments, we ask: \emph{how does token assignment during new-task training lead to routing confusion?}

\par
We investigate how different tokens from new-task data are assigned to experts during training and how this finally influences routing and performance on both tasks in testing. 
To closely examine the token–router dynamics when training newly added experts and router parameters, we conduct a controlled two-task experiment at the second incremental step (Fig.~\ref{fig:analyses}). 
When the second (\ie, new) task arrives, only the newly added LoRA experts and router are updated in a default way (\ie, basic IncMoELoRA; Sec.~\ref{sec:dymoe_layer}), and we measure accuracy on both new and old tasks as indicators of new-knowledge acquisition and forgetting. 
During training, each token is assigned to all experts (including old frozen ones and new learnable ones) according to routing scores $\bs=[\bs_\text{old}, \bs_\text{new}]$. 
Since routing-drift occurs with old-task tokens attracted by newly trained components, even when training only on new-task tokens, we hypothesize that different tokens exhibit varying degrees of new patterns---not all new-task tokens carry genuinely new patterns---and that freely assigning all of them to both old and new experts during training causes routing confusion between tasks. 
After investigating the token-expert assignment pattern, we dynamically categorize new-task tokens into three groups based on the relative dominance of $\bs_\text{new}$ vs.\ $\bs_\text{old}$ in their routing scores $\bs=[\bs_\text{old}, \bs_\text{new}]$: \textit{new}, \textit{old}, and \textit{ambiguous} tokens. 
We analyze how each token type influences forgetting and new-task learning (Fig.~\ref{fig:analyses}), yielding three key observations.

\begin{figure*}[!t]
\centering
\includegraphics[width=0.91\textwidth]{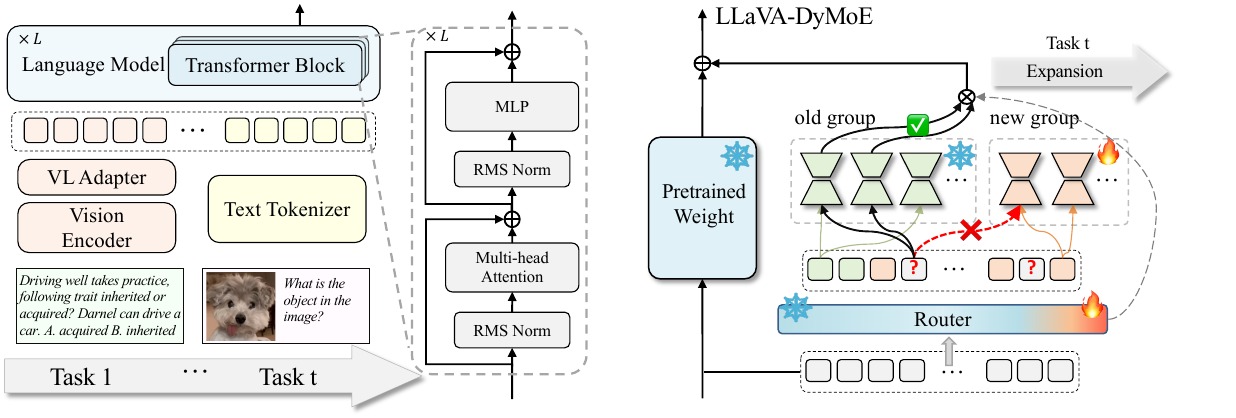}
\vspace{-0.1cm}
\caption{Overview of our LLaVA-DyMoE method, which applies a dynamic MoE with LoRA experts to each layer of the language backbone in LLaVA. It is a two-fold regularization approach designed to resolve routing-drift-induced forgetting, based on our analysis of different token types in Sec.~\ref{sec:analyze}. The right panel illustrates this high-level approach: as new tasks (Task $t$) arrive, the router and experts expand, creating a frozen ``\textcolor{Green}{old group}" and a trainable ``\textcolor{Orange}{new group}". Our Token Assignment Guidance (TAG) prevents routing-drift (red dashed arrow) by directing tokens to appropriate expert–router groups, complemented by our Routing Score Regularization (RSR) that encourage exclusive token-to-group routing and new-expert specialization. Our method regularizes the router behavior during training and imposes no constraints at inference, allowing seamless combination with other continual learning methods.}\label{fig:main}
\vspace{-0.2cm}
\end{figure*}

\par
\noindent\textbf{Observation 1}: New tokens (with high affinity to the new expert group) primarily drive new-task knowledge acquisition and cause less forgetting. As shown in Fig.~\ref{fig:analyses}(a), training only on new tokens yields strong new-task performance with minimal forgetting, as these tokens carry patterns distinct from old tasks and are naturally routed to newly added experts, leaving the old router policy uncorrupted.

\noindent\textbf{Observation 2}: Old tokens contribute less to new-task learning. Masking them from accessing newly added parameters yields similar new-task performance and forgetting as the baseline (Fig.~\ref{fig:analyses}(b)), suggesting they are best handled by old frozen experts and do not need to contribute to new-task learning. When assigned small but non-negligible weight toward new experts (by an under-optimized router), this inadvertently biases the new router toward old-task patterns, causing routing-drift despite limited learning value.

\noindent\textbf{Observation 3}: Ambiguous tokens offer minimal new-task learning benefit while posing a direct forgetting risk. Identified by their small affinity difference between old and new expert groups, these tokens capture ambiguous patterns across tasks.
Their ambiguity makes them particularly difficult to handle correctly. As shown in Fig.~\ref{fig:analyses}(c), training solely on these tokens neither improves new-task acquisition nor preserves old-task performance, confirming their minimal learning value and direct forgetting risk.

\par
These controlled experiments reveal how different token types affect new-task learning and contribute to routing-drift, exposing the link between the plasticity--stability dilemma in CL and the \textit{token's dilemma}: the inherent assignment ambiguity and trade-off between learning new tasks and inducing routing-drift. Building on this insight, we design regularization strategies for DyMoE that identify token types and guide their assignment during training, enabling us to leverage all tokens while mitigating routing-drift-induced forgetting.

\subsection{Drift-Aware Token Assignment Regularization for Alleviating Forgetting}\label{sec:meth-DATAR}
To resolve routing-drift-induced forgetting in DyMoE for MCIT, we design a two-fold regularization approach in our proposed {LLaVA-DyMoE}. 
As analyzed in Sec.~\ref{sec:analyze}, different new-task tokens from new component training affect new-task learning and old-task forgetting differently. Motivated by this, our proposed regularization guides token routing between old frozen and newly added experts during training to mitigate routing-drift, which relies solely on tokens' routing scores without additional assumptions. The regularization operates on intermediate token representations across all MoE layers.
The two-fold regularization comprises Token Assignment Guidance (TAG) and Routing Score Regularization (RSR). 
TAG identifies token types from their routing scores and guides their assignment by adjusting routing scores during training, shaping the router to avoid drift. It directly tackles the tokens' dilemma and specifically handles ambiguous tokens. 
As a complementary soft regularization, RSR directly regularizes the routing score values to enforce discrepancy and specialty.

\subsubsection{Token Assignment Guidance (TAG)}\label{sec:tag}
During training, router behavior in MoE is iteratively updated through backpropagation, and the token assignments made by an under-optimized router directly affect the subsequent learning of both experts and routers, influencing whether the model develops the desired expert specializations and routing patterns~\cite{xie2023moec,wei2025routing,yang2025solving}. 
As routers and experts are jointly trained, under-optimized routing weights may generate misleading gradients through token-expert assignment, contaminating the training of both. 
Our analysis in Sec.~\ref{sec:analyze} shows that different tokens influence training differently w.r.t. old and new expert groups: new tokens carry clear new patterns and route naturally to new experts; old tokens gravitate toward old experts but their residual affinity for new experts should be suppressed to prevent routing corruption; ambiguous tokens exhibit uncertain routing between both groups and require careful handling. TAG dynamically identifies token types via their routing score ambiguity w.r.t.\ old and new expert groups and guides their assignment during training to mitigate routing-drift.

\noindent\textbf{Token assignment ambiguity w.r.t. router-expert group.}
In LLaVA-DyMoE, when a new task $t$ arrives in MCIT, we add new experts together with their router parameters, resulting in two router–expert groups: the old group $\mcS_{t-1}$ (also denoted $\mcS_{t,\text{old}}$) and the new group $\mcS_{t,\text{new}}$.
For the representation of a given token at any layer, we denote the router logits over all experts as
 $\bs_t=[\bs_{t-1}; \bs_{t, \text{new}}] \in \mbbR^{|\mcE_{t}|}$. 
We extract the confidence score from each expert group by taking the maximum logit within that group:
\begin{equation}
c_{\text{old}} = \max(\bs_{t-1}),
\quad
c_{\text{new}} = \max(\bs_{t, \text{new}}).
\end{equation}
We quantify token assignment \emph{ambiguity} by the relative difference between the two group-wise confidence scores:
  \begin{equation}
D_{\text{rel}} = \frac{|c_{\text{new}} - c_{\text{old}}|}{\max(|c_{\text{new}}|, |c_{\text{old}}|) + \epsilon} ,
\end{equation}  
where $\epsilon$ is a small constant (\eg, ${1e-9}$) for numerical stability. $D_{\text{rel}}$ characterizes the token type. By introducing the ambiguity hyperparameter $\tau$, a token (\ie, intermediate representation) is identified as ambiguous if $D_{\text{rel}}\leq \tau$, indicating no clear routing preference between 
old and new expert groups.

\noindent\textbf{Token assignment guidance.}
The TAG mechanism routes a token to new expert group $\mcS_{t, \text{new}}$ only if it meets two conditions simultaneously: (1) it is not ambiguous ($D_{\text{rel}} > \tau$), and (2) it is new-dominant ($c_{\text{new}} > c_{\text{old}}$).

We formalize this decision with a binary mask $M_{\text{new}} \in \{0, 1\}$:
$M_{\text{new}} = \mathbbm{1}\left( (c_{\text{new}} > c_{\text{old}}) \land (D_{\text{rel}} > \tau) \right)$,
where $\mathbbm{1}()$ is the indicator function, $M_{\text{old}} = 1 - M_{\text{new}}$. This ensures that any token which is old-dominant ($c_{\text{old}} \ge c_{\text{new}}$) or ambiguous ($D_{\text{rel}} \le \tau$) is automatically assigned to the \emph{safe} old expert group. TAG applies the mask to produce the final logits $\bs'_t \in \mbbR^{|\mcE_{t}|}$ that are passed to the Softmax function. 
For expert $i$, the final routing score $s'_{t,i}$ is defined as:
\begin{equation}
s'_{t,i} =
\begin{cases}
s_{t,i}, & \text{if } m_{t,i} = 1, \\
-\infty, & \text{otherwise},
\end{cases}
\end{equation}
where $m_{t,i} = \mathbbm{1}(i \in \mcS_{t-1})\, M_{\text{old}}
        + \mathbbm{1}(i \in \mcS_{t,\text{new}})\, M_{\text{new}}.$
TAG routes low-ambiguity tokens according to their inherent-pattern-decided routing preference: new tokens to new experts to promote new-task learning, and old tokens to old experts with their residual affinity for new experts suppressed. 
Considering their low benefit for learning new knowledge (Sec.~\ref{sec:analyze}), high-ambiguity tokens are routed safely to old experts to prevent potential forgetting, when no additional prior knowledge is given.

\subsubsection{Routing Score Regularization (RSR)}
\label{sec:loss}
As a complement to TAG, we introduce Routing Score Regularization (RSR) to directly regularize the routing score weights. RSR comprises two terms: (1) an exclusivity loss ($\mathcal{L}_{\text{exc}}$) that enforces clean separation between old and new expert groups; and (2) a specialization loss ($\mathcal{L}_{\text{spe}}$) that promotes new-expert utilization and specialization to enhance new-task learning.

Given a token representation at a specific layer, we define the collective gate output, \ie, the total routing probability mass assigned to each expert group, in terms of routing weights $w_i$ (from Eq.~(\ref{eq:moelora})) as:
\begin{equation}
g_{\text{old}}=\sum\nolimits_{i\in\mcS_{t-1}} w_i,
\quad
g_{\text{new}}=\sum\nolimits_{i\in\mcS_{t,\text{new}}} w_i.
\end{equation}

\noindent\textbf{Exclusivity loss.}
This loss enforces clean routing separation by preventing a token from strongly activating both expert groups simultaneously, directly working on the routing scores. Minimizing the product of their collective gate outputs encourages exclusive routing to one group, reinforcing the conditional routing decision of TAG:
\begin{equation}\label{eq:exc}
\mathcal{L}_{\text{exc}}=g_{\text{old}}\,g_{\text{new}}.
\end{equation}

\noindent\textbf{Specialization loss.}
Complementing TAG and $\mathcal{L}_\text{exc}$ that mitigate routing-drift and forgetting for \emph{stability}, we introduce the specialization loss $\mathcal{L}_\text{spe}$ to promote and balance \emph{plasticity} by encouraging higher routing weight toward new experts. 
We first define a soft target $y$ which is close to 1 if no old experts are selected by the router (\ie, $\tilde{g}_{\text{old}}=\max\{w_i\}_{i\in\mcS_{t-1}}$ is close to zero): $y\triangleq 1 - \tilde{g}_{\text{old}}$. Relying on $\tilde{g}_{\text{old}}$, $y$ approaches $1$, reflecting the 
varying activity level of old experts. 
$\mathcal{L}_{\text{spe}}$ is formulated as a BCE loss between the collective new expert routing weight $g_{\text{new}}$ and target $y$ to encourage usage of new experts:
\begin{equation}\label{eq:spe}
\mathcal{L}_{\text{spe}} = -\,y\log g_{\text{new}} -(1-y)\log\!\big(1-g_{\text{new}}\big).
\end{equation} 
$\mathcal{L}_{\text{spe}}$ works in synergy with $\mathcal{L}_{\text{exc}}$ and TAG to balance new-task learning and forgetting mitigation.

\subsection{Training of LLaVA-DyMoE}
\noindent\textbf{Training objectives.} Our total training objective is a weighted combination of the primary task-learning loss, a standard auxiliary load balancing loss for MoE, and our proposed routing regularization terms:
\begin{equation}
\mathcal{L} = \mathcal{L}_\text{NTP} + \lambda \mathcal{L}_\text{aux} + \alpha (\mathcal{L}_\text{exc} + \mathcal{L}_\text{spe}) ,
\end{equation}
where $\mathcal{L}_\text{NTP}$ is the standard instruction-tuning loss (\eg, autoregressive cross-entropy), and $\lambda$ and $\alpha$ are scalar hyperparameters. We adopt the standard auxiliary load balancing loss ($\mathcal{L}_\text{aux}$)~\cite{fedus2022switch} to ensure balanced utilization, applied to the set of newly added experts. We use a unified hyperparameter $\alpha$ to control the contributions of the two complementary regularization losses, $\mathcal{L}_\text{exc}$ and $\mathcal{L}_\text{spe}$.

\begin{table*}[!t]
\caption{Comparison with continual learning models on the CoIN benchmark.}
\vspace{-0.1cm}
\renewcommand{\arraystretch}{1.0}
\label{tab:main}
\small
\centering
\resizebox{0.96\linewidth}{!}{
\begin{tabular}{lcccccccc|ccc}
\toprule

\multirow{2}{*}{\textbf{Method}} & \multicolumn{8}{c|}{\textbf{Accuracy on Each Task (\%)}} & \multicolumn{3}{c}{\textbf{Aggregate Results (\%)}} \\
\cmidrule(lr){2-9} \cmidrule(lr){10-12}
& SQA & VQA\textsuperscript{Text} & ImgNet & GQA & VizWiz & REF & VQA\textsuperscript{v2} & VQA\textsuperscript{OCR}  & MFN$\uparrow$ & MAA$\uparrow$ & BWT$\uparrow$  \\ 
\midrule
{LoRA} &52.56&48.12&39.27&44.47 &37.46 &1.22 &56.10 &55.11 &41.79 &43.99 &-23.12\\
{MoELoRA~\cite{chen2024coin}} &72.01&46.89&44.75&42.79&28.22&3.31&55.74&57.72   &43.93 &43.92 & -22.18 \\
{EWC~\cite{schwarz2018progress}} &59.11&47.21&39.88&45.12&35.33&2.72&56.29&41.21 &40.86 &43.75 &-21.76 \\
{LWF~\cite{li2017learning}}&62.32&48.66&51.45&45.84&43.76&0.24&54.96&44.63& 43.98& 44.89&-19.69 \\
{IncLoRA} &73.33&44.32&54.59&44.07&25.93&4.49&54.91&58.55 &45.02 &43.12 &-23.21\\
{O-LoRA~\cite{wang2023orthogonal}} &75.61&49.98&78.24&44.18&30.70&4.66&55.51&57.37  &49.53 &46.65 &-17.54\\
{IncMoELoRA} &68.43&50.31&68.42&47.97&39.46&4.56&57.31&60.95&49.68&49.50&-16.67\\
\midrule
LLaVA-DyMoE (Ours) &\textbf{76.25}&\textbf{53.86}&\textbf{95.80}&\textbf{48.40}&\textbf{52.35}&\textbf{9.25}&\textbf{58.30}&\textbf{62.00} & \textbf{57.03}& \textbf{57.70}& \textbf{-4.67}\\
\bottomrule
\end{tabular}
}
\end{table*}

\noindent \textbf{Integration with other methods.} 
Our {LLaVA-DyMoE}, which focuses on rectifying micro-level token routing-drift, is inherently orthogonal to and compatible with other macro-level MCIT paradigms. It is also compatible with data-based approaches~\cite{rebuffi2017icarl,chen2025sefe,yu2025progressive}, as LLaVA-DyMoE enhances the router's robustness in handling the mixed stream of old and new data, regardless of its source. Our method can also be seamlessly integrated into architectures that employ task-level routing methods~\cite{yu2025progressive,guo2025hide,zhao2025llava}. These methods first decide which group of experts to activate at a task level, while our LLaVA-DyMoE then optimizes the token assignment within that activated group, mitigating the intra-group routing drifts we identified. These combinations offer the potential for further enhanced performance.

\begin{table*}[t!]
  \centering
  \begin{minipage}{0.32\textwidth}
    \centering
    \small
    \centering
    \caption{Ablations on main components.}
    \vspace{-0.1cm}
    \resizebox{\linewidth}{!}{
    \small
    \begin{tabular}{l ccc}
    \toprule
    \multirow{2}{*}{\centering Configuration} & \multicolumn{3}{c}{\textbf{Aggregate Results (\%)}} \\
    \cmidrule(lr){2-4}
     & MFN$\uparrow$ & MAA$\uparrow$ & BWT$\uparrow$ \\
    \midrule
    IncMoELoRA & 49.68 & 49.50 & -16.67 \\
     + $\mathcal{L}_\text{aux}$ & 50.76 & 51.17 & -15.44 \\
     + TAG & 54.44 & 52.18 & -7.04 \\
     + $\mathcal{L}_\text{exc}$ & 55.18 & 54.38 & -6.83 \\
     + $\mathcal{L}_\text{spe}$ {(Ours)} & {57.03} & {57.70} & {-4.67} \\
    \bottomrule
    \end{tabular}
    }
\label{tab:ab}
  \end{minipage}
  \hfill
  \begin{minipage}{0.31\textwidth}
    \centering
\caption{Ablations on ambiguity threshold.}
  \vspace{-0.1cm}
  \small
  \begin{tabular}{c ccc}
    \toprule
    \multicolumn{1}{c}{\multirow{2}{*}{ $\tau$}} & \multicolumn{3}{c}{\textbf{Aggregate Results (\%)}} \\
    \cmidrule(lr){2-4}
     & MFN$\uparrow$ & MAA$\uparrow$ & BWT$\uparrow$ \\
    \midrule
    10\% &56.87  &57.23  &-4.94  \\
    20\% &57.03  &57.70  &-4.67  \\
    30\% &56.27  &55.65  &-5.21 \\
    50\% &55.32  &53.51  &-5.54 \\
    \bottomrule
  \end{tabular}
  \label{tab:ab_ratio}
  \end{minipage}
  \hfill
  \begin{minipage}{0.3\textwidth}
    \centering
  \caption{Ablations on loss weights $\alpha$.}
  \vspace{-0.1cm}
  \small
  \begin{tabular}{c ccc}
    \toprule
    \multicolumn{1}{c}{\multirow{2}{*}{ $\alpha$}} & \multicolumn{3}{c}{\textbf{Aggregate Results (\%)}} \\
    \cmidrule(lr){2-4}
     & MFN$\uparrow$ & MAA$\uparrow$ & BWT$\uparrow$ \\
    \midrule
    1e-2 &55.43&57.73&-5.81  \\
    5e-3 &56.87&57.50&-5.32  \\
    1e-3 &57.03  &57.70  &-4.67 \\
    5e-4 &56.32&57.63&-4.94 \\
    \bottomrule
  \end{tabular}
  \label{tab:ab_weight}
  \end{minipage}
  \vspace{-0.1cm}
\end{table*}

\section{Experiments} \label{sec:exper}
\subsection{Experimental Setup}
\noindent\textbf{Datasets.} We evaluate our method on the CoIN~\cite{chen2024coin} benchmark, which encompasses a series of eight VQA tasks. These tasks include ScienceQA (SQA)~\cite{lu2022learn}, TextVQA~\cite{singh2019towards}, ImageNet~\cite{imagenet15russakovsky}, GQA~\cite{hudson2019gqa}, VizWiz~\cite{gurari2018vizwiz}, RefCOCO (REF)~\cite{kazemzadeh-etal-2014-referitgame}, VQAv2~\cite{goyal2017making}, and OCR-VQA~\cite{mishra2019ocr}. Each task varies in terms of the number of data samples, stylistic features, and domain characteristics. The training set comprises a total of 569k samples, while the testing set contains 261k samples.

\noindent\textbf{Evaluation metrics.}
We adopt the metrics introduced in CoIN~\cite{chen2024coin,chen2025sefe}, which measure the discrepancy between the model’s output and the ground truth. Specifically, we report: (1) Mean Final Accuracy $\text{MFN} = \frac{1}{T}\sum_{i=1}^{T} A_{T,i}$, assessing the average accuracy across all tasks after the complete incremental training sequence, where $A_{T,i}$ represents the accuracy on task $i$ after learning on the final task $T$; (2) Mean Average Accuracy $\text{MAA} = \frac{1}{T}\sum_{j=1}^{T} \frac{1}{j} \sum_{i=1}^{j} (A_{j,i})$, representing the mean of the average accuracies on all learned tasks after each incremental training step; and (3) Backward Transfer $\text{BWT} = \frac{1}{T}\sum_{i=1}^{T} (A_{T,i} - A_{i,i})$, assessing the degree of forgetting.

\noindent\textbf{Implementation details.}
In our experiments, we utilize the pre-trained, instruction-untuned LLaVA-v1.5-7B~\cite{liu2023visual} as the backbone model for continual learning. The model integrates Vicuna~\cite{chiang2023vicuna} as its language backbone and a pre-trained CLIP ViT-L/14 visual encoder~\cite{radford2021learning} to extract visual embeddings.
Only the newly added modules are trainable, while the other components remain frozen throughout the continual learning process.
Please refer to the Appendix for details on the network architectures, hyperparameters, and implementation settings.

\begin{table*}[!t]
\caption{LLaVA-DyMoE is compatible with data-based continual learning strategies.}
\label{tab:variant}
\vspace{-0.1cm}
\renewcommand{\arraystretch}{1.0}
\small
\centering
\resizebox{0.96\linewidth}{!}{
\begin{tabular}{llcccccccc|ccc}

\toprule
&\multirow{2}{*}{\textbf{Method}} & \multicolumn{8}{c|}{\textbf{Accuracy on Each Task (\%)}} & \multicolumn{3}{c}{\textbf{Aggregate Results (\%)}} \\
\cmidrule(lr){3-10} \cmidrule(lr){11-13}
&& SQA & VQA\textsuperscript{Text} & ImgNet & GQA & VizWiz & REF & VQA\textsuperscript{v2} & VQA\textsuperscript{OCR}  & MFN$\uparrow$ & MAA$\uparrow$ & BWT$\uparrow$  \\ 
\midrule
\multicolumn{1}{c}{\multirow{2}{*}{{None}}} &{MoELoRA~\cite{chen2024coin}} &72.01&46.89&44.75&42.79&28.22&3.31&55.74&57.72   &43.93 &43.92 & -22.18 \\
&{O-LoRA~\cite{wang2023orthogonal}} &75.61&49.98&78.24&44.18&30.70&4.66&55.51&57.37  &49.53 &46.65 &-17.54\\
\midrule
\multirow{2}{*}{{ + ASD~\cite{chen2025sefe}}}&{SEFE~\cite{chen2025sefe}}  & 75.35 & 58.66 & 83.10 & 54.25 & 48.85 & 16.75 & 65.35 & 66.25  & {58.57} & {63.04} & {-10.45}   \\
&{LLaVA-DyMoE (Ours)} & 74.60&55.24&93.80&53.45&55.00&25.50&63.95&62.85&60.55&62.26&-4.75\\
\midrule
\multirow{2}{*}{{ + Replay}}&{ProgLoRA~\cite{yu2025progressive}} & 74.84 &51.83 &83.90 &49.93 &53.87 &31.19 &62.71 &64.44  &59.09 &62.38 &-6.59\\
&{LLaVA-DyMoE (Ours)}&75.55&56.88&96.50&54.75&55.90&29.15&63.65&64.25&62.08&61.93&-1.55 \\
\bottomrule
\end{tabular}
}
\vspace{-0.1cm}
\end{table*}

\subsection{Main Results}
We evaluate the proposed method on the CoIN~\cite{chen2024coin} benchmark, as presented in Table \ref{tab:main}. Previous methods that continually train a static model without expanding new parameters for new tasks, e.g., LoRA, MoELoRA~\cite{chen2024coin}, EWC~\cite{schwarz2018progress}, and LWF~\cite{li2017learning}, show significant forgetting and inferior accuracy as the knowledge obtained from previous tasks gets repeatedly overwritten.

To mitigate such overwriting, new modules are expanded to explicitly capture this new knowledge. We conduct baseline experiments with IncLoRA (adding new LoRA modules for each task), and O-LoRA~\cite{wang2023orthogonal}, which further adds orthogonal regularization to the LoRA weight matrix to help mitigate forgetting by regularizing parameter updates in orthogonal directions. To ensure that LoRA experts learned from previous tasks can be reused while new task knowledge can be absorbed without excessive overwriting, we further implement the per-task expansion baseline in the MoE framework (IncMoELoRA), where new LoRA experts are added for each new task. This baseline exhibits comparable to or better performance than O-LoRA without any regularization techniques. However, it suffers from ambiguous token-expert assignment (see Fig.~\ref{fig:analyses}), where ambiguous cross-task tokens activate and train newly added experts, largely hindering old expert utilization on these tokens.

Our proposed method significantly outperforms these methods through our token-level routing regularization. This enforces the assignment of ambiguous cross-task tokens to old, frozen experts, bringing improvements of 7.35\%, 8.20\%, and 12.00\% on MFN, MAA, and BWT, respectively.
We provide qualitative studies in the Appendix.

\subsection{Ablation Study}
\noindent\textbf{The effect of main components.}
In Table \ref{tab:ab}, we ablate the effects of each component of our method. Starting from our implemented baseline (IncMoELoRA), adopting a standard auxiliary load balancing loss ($\mathcal{L}_\text{aux}$) yields a slight performance improvement by balancing the expert usage for the new incoming task. Notably, the proposed TAG module, which applies group-wise routing guidance based on token assignment ambiguity, significantly improves the final accuracy and mitigates forgetting. This is because TAG prevents ambiguous tokens from contributing to the learning of the new task (aligned with our observations in Sec.~\ref{sec:analyze}). Building on the TAG mechanism, we further regularize the raw router logits. Adding $\mathcal{L}_\text{exc}$, which regularizes the activations of ambiguous tokens on new task routers, brings a further improvement. Finally, to ensure effective learning on the new task, we enforce the utilization of new task routers on (potentially) new task-specific tokens through $\mathcal{L}_\text{spe}$. This component largely improves new task learning while maintaining low forgetting, bringing the full model's performance to the state-of-the-art level.

\noindent\textbf{The effect of ambiguity threshold $\tau$ for token assignment routing guidance.}
We ablate the effects of different ambiguity thresholds in Table \ref{tab:ab_ratio}, where we sweep the threshold over $\{10\%, 20\%, 30\%, 50\%\}$. The ambiguity threshold controls the token-expert assignment to the old and new groups of experts. A higher ambiguity threshold encourages ambiguous tokens to be assigned to frozen old experts, which helps mitigate forgetting (as these experts do not contribute to the learning of the new router and experts) but could limit learning on new tasks, and vice versa. Through our experiments, we find that controlling the ambiguity threshold between $10\%$ and $20\%$ brings the best trade-off between learning new tasks and forgetting old tasks. Using a larger ambiguity threshold hinders effective learning on the new task, as shown by the decreased MFN and MAA.

\noindent\textbf{The effect of token assignment regularization weighting factor $\alpha$.}
We conduct an ablation study on the hyperparameter $\alpha$, which controls the unified weighting of our proposed regularization losses ($\mathcal{L}_\text{exc}$ and $\mathcal{L}_\text{spe}$). As shown in Table \ref{tab:ab_weight}, we evaluate model performance by sweeping $\alpha$ across the values $\{\text{1e-2}, \text{5e-3}, \text{1e-3}, \text{5e-4}\}$. The results indicate that our proposed method is relatively robust to the choice of this weighting factor, and we adopt $\alpha = \text{1e-3}$ by default.

\subsection{Discussion}
\noindent\textbf{Study of complementary data-based techniques.} Our proposed method is orthogonal to and compatible with data-based techniques. As shown in Table~\ref{tab:variant}, incorporating these techniques with our method yields consistent improvements across metrics, demonstrating their complementary nature. We first compare against SEFE~\cite{chen2025sefe}, which adopts the ASD data paradigm and achieves a BWT of -10.45\%. By complementing this same ASD paradigm with our proposed LLaVA-DyMoE, forgetting is significantly mitigated, improving the BWT to -4.75\%. We evaluate against replay-based methods. When combined with a standard replay buffer~\cite{rolnick2019experience} (using a buffer size comparable to ProgLoRA~\cite{yu2025progressive}), our approach again outperforms the competitor: LLaVA-DyMoE with a small replay buffer achieves a BWT of -1.55\%, substantially better than ProgLoRA's -6.59\%, while also attaining a higher mean final accuracy.

\section{Conclusion}
We analyze routing-drift in dynamic MoE expansion and trace its cause to the token level: ambiguous and old tokens in new-task data offer minimal learning value yet corrupt the router's policy for old tasks when left unguided. We propose LLaVA-DyMoE, comprising a Token Assignment Guidance mechanism that detects and redirects high-drift tokens, and complementary routing losses that enforce expert-group separation. Experiments show that LLaVA-DyMoE effectively mitigates routing-drift. As it targets the inherent routing mechanism, it is complementary to existing MCIT paradigms for further gains.
\par
\noindent\textbf{Limitations and future work.} Future work could further investigate the scalability of our approach on larger-scale models and in more realistic scenarios.

\newpage
\section*{Acknowledgments}
This work was partially supported by the ARC DECRA Fellowship (DE230101591), the ARC Discovery Project Grant (DP260103379), and the NVIDIA Academic Grant Program.

{
    \small
    \bibliographystyle{ieeenat_fullname}
    \bibliography{main}
}

\maketitlesupplementary
\appendix

\section{Additional Experiment Details}
\subsection{More Implementation Details}
In our experiments, we utilize the pre-trained, instruction-untuned LLaVA-v1.5~\cite{liu2023visual} as the backbone model for continual learning. The model employs Vicuna~\cite{chiang2023vicuna} as its language backbone and a pre-trained CLIP ViT-L/14 visual encoder~\cite{radford2021learning} for visual feature extraction. During continual learning, only the newly added modules are trainable, while all other components remain frozen. In the default setting, 16 rank-4 LoRA experts are added when each new task starts, and $K=16$ is applied for top-K routing over all trained experts. Different configurations are evaluated in ablation studies. The same configurations are also applied to the baseline model IncMoELoRA in experiments. 
All experiments are conducted on a compute node equipped with four NVIDIA H100 GPUs. Following the official LLaVA-v1.5 configuration, we adopt a global batch size of 128 and a learning rate of $2 \times 10^{-4}$. 
We set the warmup ratio to 0.03 and use the AdamW optimizer for training.
The model is trained in PyTorch with BF16 precision and DeepSpeed ZeRO-2. We set the weights of the load balancing, exclusivity, and specialization losses to $1 \times 10^{-3}$.
For all compared methods, we follow the default configurations from their original papers.
Other remaining settings are consistent with those specified for LLaVA-v1.5~\cite{liu2023visual}.

\subsection{Details of Datasets}
The eight tasks in the CoIN~\cite{chen2024coin} benchmark are as follows:

\noindent\textbf{ScienceQA (SQA)}~\cite{lu2022learn} is a multimodal science question-answering dataset designed to assess models’ reasoning over integrated visual and textual information. The training set contains 12,726 samples (6,218 image–text and 6,508 text-only), and the test set includes 4,241 samples (2,017 image–text and 2,224 text-only).

\noindent\textbf{TextVQA}~\cite{singh2019towards} focuses on text recognition within visual question-answering. It features real-world images with diverse textual content. The training set includes 34,602 image–text samples, and the test set comprises 5,000 image–text samples.

\noindent\textbf{ImageNet}~\cite{imagenet15russakovsky} is a large-scale benchmark for image classification. The training set contains 129,833 image–text samples, and the test set includes 5,050 image–text samples.

\noindent\textbf{GQA}~\cite{hudson2019gqa} emphasizes real-world visual reasoning, requiring understanding of object relationships and multi-step inference based on both synthetic and real images with scene graphs. The training and test sets include 72,140 and 12,578 image–text samples, respectively.

\noindent\textbf{VizWiz}~\cite{gurari2018vizwiz} is designed for visual question-answering in assistive contexts for visually impaired users. It provides 20,523 training samples and 4,319 test samples, all in the image–text modality.

\noindent\textbf{Grounding (Ref)}~\cite{kazemzadeh-etal-2014-referitgame} evaluates grounding of natural-language expressions in images. It contains image–text pairs requiring models to predict bounding boxes aligned with textual descriptions. The training set includes 55,885 samples, and the test set includes 30,969 samples.

\noindent\textbf{VQAv2}~\cite{goyal2017making}, a visual question-answering benchmark, features balanced answer distributions and broad topical coverage. It provides 82,783 training samples and 214,354 test samples, all in the image–text modality.

\noindent\textbf{OCRVQA}~\cite{mishra2019ocr} integrates OCR with visual question-answering to assess models’ ability to extract and reason over textual content in images. The dataset includes 165,348 training samples and 99,926 test samples, all image–text.

\begin{table}[t]
  \renewcommand{\arraystretch}{1.0}
  \centering
  \caption{Performance under varying task orders.}
  \begin{tabular}{c ccc}
    \toprule
    \multirow{2}{*}{ \makecell[c]{Task \\Order}} & \multicolumn{3}{c}{\textbf{Aggregate Results (\%)}} \\
    \cmidrule(lr){2-4}
     & MFN$\uparrow$ & MAA$\uparrow$ & BWT$\uparrow$ \\
    \midrule
    Origin &57.03  &57.70  &-4.67  \\
    Reverse &56.67 &57.34 & -4.71  \\
    Alphabet &56.44 & 56.98 & -4.92 \\
    \bottomrule
  \end{tabular}
  \label{tab:order}
\end{table}

\section{More Experiments}
\subsection{Experiments on Different Task Orders} 
To assess order sensitivity, we trained our method under multiple task orderings on the eight CoIN datasets. We conduct experiments using three task orderings: ``Origin'', the original task ordering proposed in the CoIN benchmark; ``Reverse'', the reversed version of the original ordering; and ``Alphabet'', where tasks are ordered alphabetically. As summarized in Table~\ref{tab:order}, our method exhibits excellent stability across different task orderings; the amount of forgetting remains consistently low with minimal variation across orderings. Without techniques like experience replay, our proposed token-level expert assignment regularization within this incremental MoE with LoRA approach consistently learns new task knowledge with minimal forgetting, regardless of the task ordering. This is because our method effectively prevents ambiguous tokens from contributing to the learning of new routers and experts, ensuring that effective new knowledge is absorbed into the new experts without any assumptions about the incoming order of the data.

\begin{table}[t]
  \renewcommand{\arraystretch}{1.0}
  \centering
  \caption{Performance under distinct training instruction types.}
  \begin{tabular}{c ccc}
    \toprule
    \multirow{2}{*}{\centering Instruction} & \multicolumn{3}{c}{\textbf{Aggregate Results (\%)}} \\
    \cmidrule(lr){2-4}
     & MFN$\uparrow$ & MAA$\uparrow$ & BWT$\uparrow$ \\
    \midrule
    Origin &57.03  &57.70  &-4.67  \\
    Diverse &56.90 & 57.44 &-5.12  \\
    10Type & 56.77 & 57.25 &-4.87\\
    \bottomrule
  \end{tabular}
  \label{tab:type}
\end{table}

\subsection{Experiments on Distinct Training Instruction Types} 
To validate the reliability of our proposed method against distinct instruction templates, we conduct experiments with different template types, reported in Table~\ref{tab:type}. 
There are three types of instruction templates in the CoIN benchmark~\cite{chen2024coin}.
Following the default setting, the experiments in the main paper are based on the ``Origin'' type. We further conduct experiments with the other two types of instruction templates, Diverse and 10Type, in~\cite{chen2024coin}: 1) Diverse: Distinct instruction templates tailored to different tasks. 2) 10Type: Randomly sampled from 10 distinct instruction templates. (Details can be found in Table~\ref{tab:diversity}.) The results show that forgetting and accuracy on all three metrics are nearly identical across instruction types, indicating the method’s stability. This result is significant, as it indicates that our method's token-level routing mechanism is not overfitting to superficial, task-specific prompt formats. 

\subsection{Ablation Studies on MoE Configurations} 
In this section, we validate the proposed method on different MoE configurations, including the top-K value, the number of experts, and the expert capacity, and show the results in Table~\ref{tab:ab_moe_topk}, \ref{tab:ab_moe_expert_number}, and \ref{tab:ab_moe_expert_capacity}. 
Experiments with baseline IncMoELoRA are conducted as a reference. 
The ablation studies demonstrate that the proposed method performs robustly across different MoE configurations and consistently delivers improvements. 

\noindent\textbf{Ablations on top-K value.} 
First, we conduct an ablation study on the top-K value, comparing top-8 and top-16 over all experts. As shown in Table~\ref{tab:ab_moe_topk}, the overall performance is relatively insensitive to the choice of top-K, and under both configurations our LLaVA-DyMoE consistently outperforms the baseline method, demonstrating the effectiveness of the proposed approach.

\begin{table}[t]
  \renewcommand{\arraystretch}{1.0}
  \centering
  \caption{Ablations on top-K value.}
  \begin{tabular}{cl ccc}
    \toprule
    \multirow{2}{*}{Top-K} & \multirow{2}{*}{ Method}& \multicolumn{3}{c}{\textbf{Aggregate Results (\%)}} \\
    \cmidrule(lr){3-5}
     && MFN$\uparrow$ & MAA$\uparrow$ & BWT$\uparrow$ \\
    \midrule
    \multirow{2}{*}{8}&IncMoELoRA &48.55 &49.87 &-16.28   \\
    &LLaVA-DyMoE&56.89 &57.71 &-5.12\\
    \midrule
    \multirow{2}{*}{16}&IncMoELoRA&49.68&49.50&-16.67  \\
    &LLaVA-DyMoE&57.03  &57.70  &-4.67\\
    \bottomrule
  \end{tabular}
  \label{tab:ab_moe_topk}
\end{table}

\begin{table}[t]
  \renewcommand{\arraystretch}{1.0}
  \centering
  \caption{Ablations on expert number.}
  \begin{tabular}{cl ccc}
    \toprule
    \multirow{2}{*}{ \makecell[c]{Expert \\ Number}} & \multirow{2}{*}{ Method}& \multicolumn{3}{c}{\textbf{Aggregate Results (\%)}} \\
    \cmidrule(lr){3-5}
     && MFN$\uparrow$ & MAA$\uparrow$ & BWT$\uparrow$ \\
    \midrule
    \multirow{2}{*}{8}&IncMoELoRA &48.39 &50.62 &-17.93   \\
    &LLaVA-DyMoE   &56.97 &58.37 &-5.78   \\
    \midrule
    \multirow{2}{*}{16}&IncMoELoRA&49.68&49.50&-16.67    \\
    &LLaVA-DyMoE &57.03  &57.70  &-4.67 \\
    \bottomrule
  \end{tabular}
  \label{tab:ab_moe_expert_number}
\end{table}

\begin{table}[t]
  \renewcommand{\arraystretch}{1.0}
  \small
  \centering
  \caption{Ablations on expert capacity.}
  \begin{tabular}{cl ccc}
    \toprule
    \multirow{2}{*}{ \makecell[c]{Expert \\ Capacity}} & \multirow{2}{*}{ Method}& \multicolumn{3}{c}{\textbf{Aggregate Results (\%)}} \\
    \cmidrule(lr){3-5}
     && MFN$\uparrow$ & MAA$\uparrow$ & BWT$\uparrow$ \\
    \midrule
    \multirow{2}{*}{1}&IncMoELoRA &48.23 &49.14 &-15.79  \\
    &LLaVA-DyMoE  &56.78 & 57.34 &-4.13\\
    \midrule
    \multirow{2}{*}{2}&IncMoELoRA &49.08 &49.25 &-16.58   \\
    &LLaVA-DyMoE  &56.91 & 57.48 &-4.62\\
    \midrule
    \multirow{2}{*}{4}&IncMoELoRA&49.68&49.50&-16.67    \\
    &LLaVA-DyMoE &57.03  &57.70  &-4.67 \\
    \bottomrule
  \end{tabular}
  \label{tab:ab_moe_expert_capacity}
\end{table}

\begin{table*}[t]
\caption{Performance across different model sizes.}
\label{tab:model_size}
\renewcommand{\arraystretch}{1.0}
\small
\centering
\resizebox{\linewidth}{!}{
\begin{tabular}{clcccccccc|ccc}
\toprule
\multirow{2}{*}{\textbf{Size}}&\multirow{2}{*}{\textbf{Method}} & \multicolumn{8}{c|}{\textbf{Accuracy on Each Task (\%)}} & \multicolumn{3}{c}{\textbf{Aggregate Results (\%)}} \\
\cmidrule(lr){3-10} \cmidrule(lr){11-13}
&& SQA & VQA\textsuperscript{Text} & ImgNet & GQA & VizWiz & REF & VQA\textsuperscript{v2} & VQA\textsuperscript{OCR}  & MFN$\uparrow$ & MAA$\uparrow$ & BWT$\uparrow$  \\ 
\midrule
\multirow{2}{*}{7B}&{IncMoELoRA} &68.43&50.31&68.42&47.97&39.46&4.56&57.31&60.95&49.68&49.50&-16.67\\
&{LLaVA-DyMoE (Ours)} &{76.25}&{53.86}&{95.80}&{48.40}&{52.35}&{9.25}&{58.30}&{62.00} & {57.03}& {57.70}& {-4.67}\\
\midrule
\multirow{2}{*}{13B}&{IncMoELoRA} &68.75&51.69&85.80&48.10&40.20&6.55&58.85&64.60& 53.07& 53.20 & -14.23\\
&{LLaVA-DyMoE (Ours)}&78.75&56.24&96.05&55.85&53.20&13.85&64.05&65.15&60.39 &61.25 &-4.64\\
\bottomrule
\end{tabular}
}
\end{table*}

\begin{table*}[h]
\caption{LLaVA-DyMoE is compatible with task-level routing methods.}
\label{tab:task_router}
\renewcommand{\arraystretch}{1.0}
\small
\centering
\resizebox{\linewidth}{!}{
\begin{tabular}{lcccccccc|ccc}
\toprule
\multirow{2}{*}{\textbf{Method}} & \multicolumn{8}{c|}{\textbf{Accuracy on Each Task (\%)}} & \multicolumn{3}{c}{\textbf{Aggregate Results (\%)}} \\
\cmidrule(lr){2-9} \cmidrule(lr){10-12}
& SQA & VQA\textsuperscript{Text} & ImgNet & GQA & VizWiz & REF & VQA\textsuperscript{v2} & VQA\textsuperscript{OCR}  & MFN$\uparrow$ & MAA$\uparrow$ & BWT$\uparrow$  \\ 
\midrule
{LLaVA-DyMoE} &{76.25}&{53.86}&{95.80}&{48.40}&{52.35}&{9.25}&{58.30}&{62.00} & {57.03}& {57.70}& {-4.67}\\
{+ Task Router} &78.18 & 53.36 & 95.63 & 54.63 & 53.92 & 24.46 & 59.54 & 60.40 & 60.02 & 60.78 &-1.73\\
\bottomrule
\end{tabular}
}
\end{table*}

\begin{table}[!h]
  \renewcommand{\arraystretch}{1.0}
  \small
  \centering
  \caption{LLaVA-DyMoE is compatible with data-based continual learning strategies. Table~\ref{tab:variant} in the main paper shows that the proposed token assignment regularization can work with replay techniques. This table shows the performance across different replay buffer sizes, with ProgLoRA~\cite{yu2025progressive} (containing replay) as a reference.}
  \begin{tabular}{c l ccc}
    \toprule
    \multirow{2}{*}{ \makecell[c]{Replay \\Size}} & \multirow{2}{*}{ Method}& \multicolumn{3}{c}{\textbf{Aggregate Results (\%)}} \\
    \cmidrule(lr){3-5}
     && MFN$\uparrow$ & MAA$\uparrow$ & BWT$\uparrow$ \\
    \midrule
    \multirow{2}{*}{ 200}&ProgLoRA  &59.09 &62.38 &-6.59  \\
    &LLaVA-DyMoE&62.08 &61.93 &-1.55\\
    \midrule
    \multirow{2}{*}{ 500} &ProgLoRA &59.14 &62.74 &-6.47 \\
     &LLaVA-DyMoE&62.55 &62.17 &-1.00\\
    \midrule
    \multirow{2}{*}{ 1000}&ProgLoRA&59.66 &63.23 &-6.21  \\
    &LLaVA-DyMoE &63.19 &62.95 &-0.64\\
    \bottomrule
  \end{tabular}
  \label{tab:ab_replay}
\end{table}

\begin{table*}[t]
\caption{Performance of LLaVA-DyMoE with expert pruning.}
\label{tab:expert_pruning}
\renewcommand{\arraystretch}{1.0}
\small
\centering
\resizebox{\linewidth}{!}{
\begin{tabular}{lcccccccc|ccc}
\toprule
\multirow{2}{*}{\textbf{Method}} & \multicolumn{8}{c|}{\textbf{Accuracy on Each Task (\%)}} & \multicolumn{3}{c}{\textbf{Aggregate Results (\%)}} \\
\cmidrule(lr){2-9} \cmidrule(lr){10-12}
& SQA & VQA\textsuperscript{Text} & ImgNet & GQA & VizWiz & REF & VQA\textsuperscript{v2} & VQA\textsuperscript{OCR}  & MFN$\uparrow$ & MAA$\uparrow$ & BWT$\uparrow$  \\ 
\midrule
{LLaVA-DyMoE} &76.25&53.86&95.80&48.40&52.35&9.25&58.30&62.00 & 57.03& 57.70& -4.67\\
{+ Pruning $1/8$} &76.04&53.91&96.10&48.16&52.51&9.27&58.21&61.79 & 57.00 & 57.62 & -4.63\\
{+ Pruning $1/4$} &75.59&53.79&95.11&47.78&52.51&9.26&57.94&61.39 &56.67 & 57.37 & -4.48\\
\bottomrule
\end{tabular}
}
\end{table*}

\begin{figure*}[th]
\centering
\begin{subfigure}{1\linewidth}
    \centering
    \includegraphics[width=\linewidth]{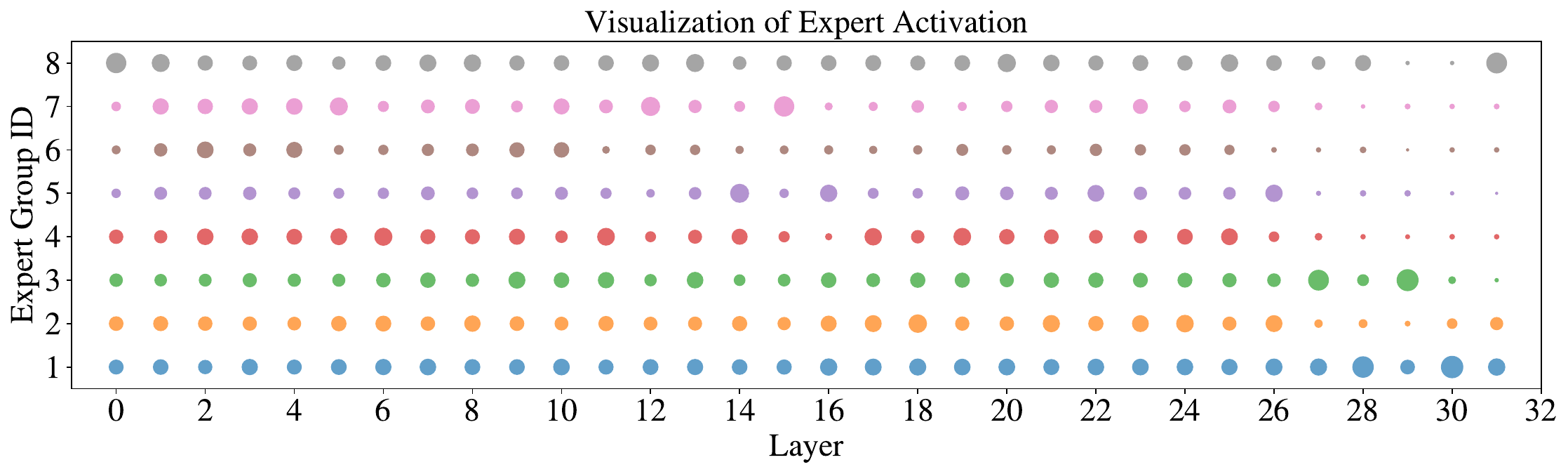}
    \subcaption{LLaVA-DyMoE after training on the $8$-th task.}
    \label{fig:moe-8}
\end{subfigure}
\begin{subfigure}{1\linewidth}
    \centering
    \includegraphics[width=\linewidth]{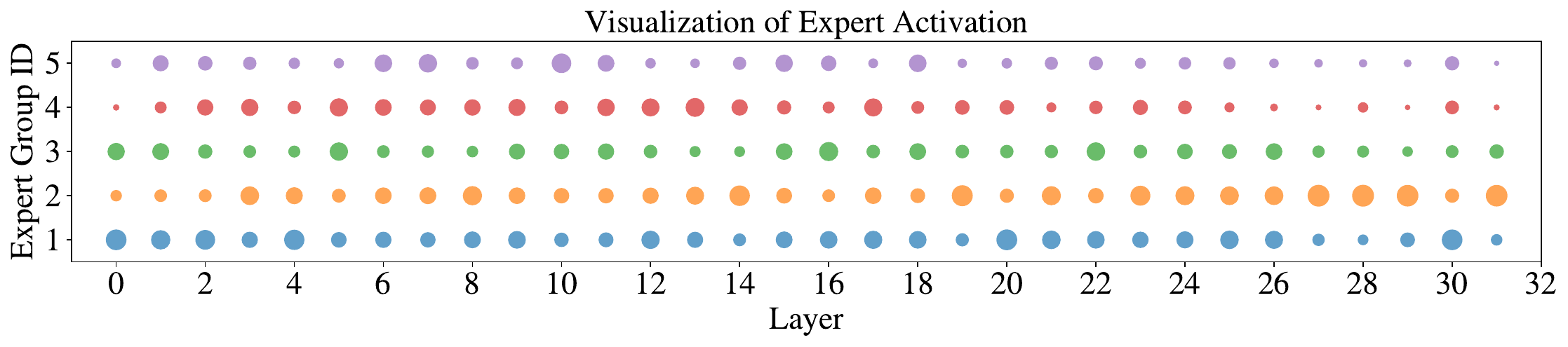}
    \subcaption{LLaVA-DyMoE after training on the $5$-th task.}
    \label{fig:moe-5}
\end{subfigure}
\caption{Layer-wise expert activation on the CoIN benchmark. Activation frequency is shown for each expert group across layers, and circle size reflects how often an expert is activated.}
\label{fig:expert_vis}
\end{figure*}

\noindent\textbf{Ablations on expert number.}
Second, we conduct an ablation study on the number of newly added experts, comparing 8 and 16 experts per task. In the 8-expert setting, we increase the parameters of each expert to maintain a comparable total capacity.
As shown in Table~\ref{tab:ab_moe_expert_number}, LLaVA-DyMoE consistently outperforms the baseline under both configurations, demonstrating its effectiveness.

\noindent\textbf{Ablations on expert capacity.}
Furthermore, we conduct an ablation study on the impact of expert capacity by varying the LoRA experts’ rank to 1, 2, and 4. As shown in Table~\ref{tab:ab_moe_expert_capacity}, models with larger expert capacity achieve better performance, and across all capacity settings LLaVA-DyMoE consistently outperforms the baseline, further demonstrating the effectiveness of the proposed approach.

Overall, these ablations show that LLaVA-DyMoE is robust to variations in the MoE configuration: it consistently outperforms the baseline across different choices of top-K, number of experts, and expert capacity. 

\subsection{Experiments with Different Backbone Sizes}
Besides the 7B model, we further validate our method using the larger 13B LLaVA backbone, as shown in Table~\ref{tab:model_size}. 
Scaling to a larger and stronger backbone model yields improved continual learning performance while preserving a low forgetting rate. LLaVA-DyMoE demonstrates robust scalability, effectively leveraging the increased capacity to achieve a higher MFN of 60.39\% while maintaining a consistently low forgetting rate (-4.64\%). This confirms that our drift-aware token assignment mechanism remains effective regardless of the underlying model size.

\subsection{Additional Results on Data-based Strategies}
The proposed drift-aware token assignment regularization in LLaVA-DyMoE is orthogonal and compatible with data-based strategies such as replay and data augmentation. By focusing on the core router training, our method can improve performance when combined with these techniques. In the main paper, we have provided the experiments in Table \ref{tab:variant}. 
In this section, we provide additional details on the results of replay techniques under different replay buffer sizes.
We compare our method, LLaVA-DyMoE equipped with a standard replay buffer~\cite{rolnick2019experience}, against ProgLoRA~\cite{yu2025progressive}.
This configuration serves as a basic replay-based variant of our dynamic MoE architecture. Following ProgLoRA, we vary the buffer size (200, 500, 1000) to match comparable replay budgets. As shown in Table~\ref{tab:ab_replay}, LLaVA-DyMoE consistently achieves competitive or better performance across different replay buffer sizes.

\subsection{Compatibility with Additional Task-specific Router}
Our LLaVA-DyMoE, which focuses on rectifying micro-level token routing drifts, is inherently orthogonal to and compatible with macro-level MCIT paradigms based on task-specific routing. In particular, our method can be seamlessly integrated into architectures that employ task-level routing strategies~\cite{yu2025progressive,guo2025hide,zhao2025llava,yu2024boosting}. These approaches first decide which group of experts to activate at the task level, while LLaVA-DyMoE then optimizes the token assignments within the selected group, mitigating the intra-group routing drifts we identified and thus providing complementary benefits.

To verify this compatibility, we equip LLaVA-DyMoE with a task-specific router.
In this setup, the task router determines which experts are activated for each task, while our dynamic MoE component can further regularize token-level routing.
As shown in Table~\ref{tab:task_router}, this combination yields improved performance over vanilla LLaVA-DyMoE, demonstrating that LLaVA-DyMoE can provide additive gains when integrated with task-specific routing methods.

\subsection{LLaVA-DyMoE with Expert Pruning}
We evaluate the performance of LLaVA-DyMoE under different MoE configurations.
In particular, we investigate expert pruning, which removes potentially unnecessary experts from the MoE.
Table~\ref{tab:expert_pruning} reports the performance of LLaVA-DyMoE after pruning either $1/8$ or $1/4$ of the experts with the lowest activation frequencies following the training of each task.
The results show that our method remains robust even with expert pruning. Note that we apply only a simple, naive pruning strategy, which leads to a slight performance drop. Although pruning is not the main focus of this work, this experiment demonstrates the potential of our proposed techniques to remain effective under more complex MoE training pipelines.

\subsection{Visualization of Expert Activation}
In Fig.~\ref{fig:expert_vis}, we present layer-wise expert activation frequencies of LLaVA-DyMoE across the eight tasks in the CoIN benchmark. For clarity, the activation frequencies of newly added experts for each task are merged into one single expert group. The routing scores are aggregated as the expert activation strength. 
The visualization shows that all experts are activated with varying strengths, exhibiting diverse utilization patterns across layers.

\begin{figure*}[th]
\centering
\includegraphics[width=0.9\textwidth]{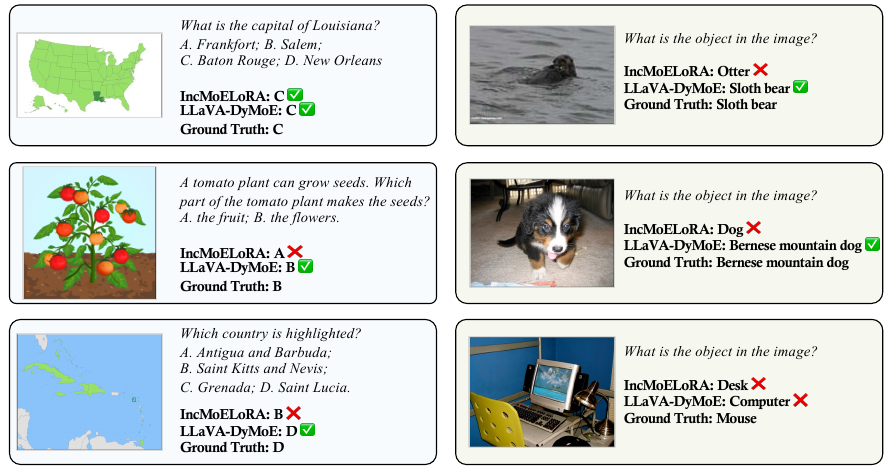}
\caption{Comparisons between baseline IncMoELoRA and LLaVA-DyMoE on cases after training on the final task. The first column shows cases from ScienceQA, the second column shows cases from ImageNet.}
\label{fig:demo}
\end{figure*}

\begin{table*}[ht]
\caption{Training time of LLaVA-DyMoE during sequential training.}
\label{tab:time}
\renewcommand{\arraystretch}{1.0}
\centering
\resizebox{0.9\linewidth}{!}{
\begin{tabular}{lccccccccc}
\toprule
\multirow{2}{*}{\textbf{Method}}  &
\multicolumn{8}{c}{\textbf{Training Time on Each Task (min)}} &\multirow{2}{*}{\textbf{Average}}\\ \cmidrule(r){2-9} 
& SQA & VQA\textsuperscript{Text} & ImgNet & GQA & VizWiz & REF & VQA\textsuperscript{v2} & VQA\textsuperscript{OCR} \\ 
\midrule
{IncMoELoRA}  & 7.4 & 13.6 & 92.4 & 123.4 & 14.9 & 99.6 & 105.0 & 137.1  &74.18\\
{LLaVA-DyMoE (Ours) }  & 7.4 & 14.6 & 95.5 & 127.6 & 15.8 & 103.2 & 109.6 &145.7 &77.43\\
\bottomrule
\end{tabular}
}
\end{table*}

\subsection{Qualitative Result Examples}
We provide a qualitative comparison in Fig.~\ref{fig:demo} by randomly sampling data from previous tasks (ScienceQA and ImageNet) after the model has finished training on the final task. As illustrated, LLaVA-DyMoE successfully retains fine-grained knowledge that the baseline often forgets. Specifically, the IncMoELoRA baseline tends to regress to coarse-grained or incorrect labels, such as simplifying a ``Bernese mountain dog'' to a generic ``Dog'', or confusing a ``Sloth bear'' with a visually similar ``Otter''. In contrast, our method accurately recalls specific species and reasoning details, such as identifying the correct biological part of a ``Tomato'' plant. While complex scenes with small objects remain challenging for both models (e.g., the ambiguous Mouse case), our approach exhibits improved knowledge retention compared to the baseline across diverse domains.

\subsection{Efficiency}
The proposed drift-aware token assignment regularization is applied during training with minor additional computations. To validate the efficiency, we report the training time of the baseline (IncMoELoRA) and our LLaVA-DyMoE in Table~\ref{tab:time}. 
Our method incurs only a small training-time overhead of 4.4\% (from 74.18 minutes to 77.43 minutes), while leaving inference efficiency unaffected.

\section{Ethical and Social Impacts}
This work advances MCIT by effectively enabling LVLMs to incrementally perform instruction tuning on new tasks while maintaining proficiency on previously learned ones. A key social benefit of our proposed LLaVA-DyMoE is its emphasis on parameter and inference efficiency. By utilizing a sparse MoE architecture, we minimize the computational energy required for long-term learning compared to dense retraining methods, aligning with the goals of Green AI. Regarding ethical considerations, we note that our model builds upon the pre-trained LLaVA backbone and standard datasets within the open-source CoIN benchmark. As with general data-driven LVLMs, our model naturally reflects the data distributions and characteristics of these foundational resources. While our current work focuses on optimizing knowledge retention and plasticity, we encourage future research to continue exploring safety alignment and fairness as integral components of the continual learning process for real-world applications.

\begin{table*}[ht]
\caption{The list of instruction templates for each task~\cite{chen2024coin}.}
\vspace{-0.2cm}
\renewcommand{\arraystretch}{1.0}
\small
\centering
\label{tab:diversity}
\resizebox{0.85\linewidth}{!}{
\begin{tabular}{cccl}
\toprule
\textbf{Task} &
\textbf{Original} &
\textbf{Diverse} &
\textbf{10Type} \\
\midrule

\multicolumn{1}{c}{\textbf{SQA}} & \makecell[c]{Answer with the option’s \\ letter from the given \\ choices directly} & \makecell[c]{Answer with the option’s \\ letter from the given \\ choices directly} & \makecell[l]{Answer with the option’s letter from the given choices directly \\ Select the correct answer from the given choices and respond with the letter of the chosen option \\ Determine the correct option from the provided choices and reply with its corresponding letter \\ Pick the correct answer from the listed options and provide the letter of the selected option \\ Identify the correct choice from the options below and respond with the letter of the correct option \\ From the given choices, choose the correct answer and respond with the letter of that choice \\ Choose the right answer from the options and respond with its letter \\ Select the correct answer from the provided options and reply with the letter associated with it \\ From the given choices, select the correct answer and reply with the letter of the chosen option \\ Identify the correct option from the choices provided and respond with the letter of the correct option \\ From the given choices, pick the correct answer and respond by indicating the letter of the correct option}  \\ \midrule

\multicolumn{1}{c}{\textbf{VQA\textsuperscript{Text}}} & \makecell[c]{Answer the question \\  using a single \\ word or phrase} & {\makecell[c]{Capture the essence of \\ your response \\ in a single word \\ or a concise phrase}} & \makecell[l]{Answer the question with just one word or a brief phrase \\ Use one word or a concise phrase to respond to the question \\ Answer using only one word or a short, descriptive phrase \\ Provide your answer in the form of a single word or a brief phrase \\ Use a single word or a short phrase to respond to the question \\ Summarize your response in one word or a concise phrase \\ Respond to the question using a single word or a brief phrase \\ Provide your answer in one word or a short, descriptive phrase \\ Answer the question with a single word or a brief, descriptive phrase \\ Capture the essence of your response in one word or a short phrase \\ Capture the essence of your response in a single word or a concise phrase} \\ \midrule

\multicolumn{1}{c}{\textbf{ImgNet}} & \makecell[c]{Answer the question \\  using a single \\ word or phrase} & {\makecell[c]{Express your answer in \\ a single word or a \\ short, descriptive phrase}} & \makecell[l]{Express your answer in a single word or a short, descriptive phrase \\ Provide your answer using a single word or a brief phrase \\ Describe the content of the image using one word or a concise phrase \\ Respond to the question with a single word or a short, descriptive phrase \\ Classify the image content using only one word or a brief phrase \\ Give your answer in the form of a single word or a concise phrase \\ Use a single word or a short phrase to categorize the image content \\ Express your answer with one word or a short, descriptive phrase \\ Identify the type of content in the image using one word or a concise phrase \\ Summarize your response in a single word or a brief phrase \\ Use one word or a short phrase to classify the content of the image} \\ \midrule

\multicolumn{1}{c}{\textbf{GQA}} & \makecell[c]{Answer the question \\  using a single \\ word or phrase} & {\makecell[c]{Respond to the question \\ briefly, using only one \\ word or a phrase}} & \makecell[l]{Respond to the question with a single word or a short phrase \\ Respond to the question using only one word or a concise phrase \\ Answer the question with a single word or a brief phrase \\ Respond with one word or a short phrase \\ Provide your answer in the form of a single word or a concise phrase \\ Respond to the question with just one word or a brief phrase \\ Answer the question using a single word or a concise phrase \\ Provide your response using only one word or a short phrase \\ Respond to the question with a single word or a brief phrase \\ Respond to the question using just one word or a concise phrase \\ Answer the question with one word or a short phrase} \\ \midrule

\multicolumn{1}{c}{\textbf{VizWiz}} & \makecell[c]{Answer the question \\  using a single \\ word or phrase} & {\makecell[c]{Provide a succinct \\ response with a single \\ word or phrase}} & \makecell[l]{Answer the question using only one word or a concise phrase \\ Respond to the question using only one word or a concise phrase \\ Respond to the question with a single word or a brief phrase \\ Provide your answer using just one word or a short phrase \\ Respond with one word or a concise phrase \\ Answer the question with just one word or a brief phrase \\ Use a single word or a short phrase to answer the question \\ Provide your answer in the form of one word or a brief phrase \\ Reply to the question using one word or a concise phrase \\ Answer with a single word or a short phrase \\ Use one word or a brief phrase to answer the question} \\ \midrule

\multicolumn{1}{c}{\textbf{REF}} & \makecell[c]{Please provide the \\ bounding box coordinate \\ of the region \\ this sentence describes} & \makecell[c]{Please provide the \\ bounding box coordinate \\ of the region \\ this sentence describes} & \makecell[l]{Identify and provide the bounding box coordinates that match the description given in this sentence \\ Extract and provide the bounding box coordinates based on the region described in the sentence \\ Please provide the bounding box coordinate of the region this sentence describes \\ Find and provide the bounding box coordinates for the region mentioned in the sentence \\ Provide the coordinates of the bounding box that correspond to the region described in the sentence \\ Give the bounding box coordinates as described in the sentence \\ Determine and provide the bounding box coordinates based on the description in the sentence \\ Identify and provide the coordinates of the bounding box described in the sentence \\ Provide the coordinates for the bounding box based on the region described in the sentence \\ Extract and provide the coordinates for the bounding box described in the sentence \\ Identify and give the coordinates of the bounding box as described by the sentence} \\ \midrule

\multicolumn{1}{c}{\textbf{VQA\textsuperscript{v2}}} & \makecell[c]{Answer the question \\  using a single \\ word or phrase} & \makecell[c]{Answer the question \\  using a single \\ word or phrase} & \makecell[l]{Answer the question using a single word or phrase \\ Answer the question with a single word or a brief phrase \\ Use one word or a short phrase to respond to the question \\ Answer the question using just one word or a concise phrase \\ Provide your answer to the question using only one word or a brief phrase \\ Respond to the question with a single word or a short phrase Use a single word or phrase to answer the question \\ Provide an answer using only one word or a brief phrase \\ Answer the question succinctly with one word or a brief phrase \\ Answer the question with just one word or a short phrase \\ Respond to the question using a single word or a concise phrase} \\ \midrule

\multicolumn{1}{c}{\textbf{VQA\textsuperscript{OCR}}} & \makecell[c]{Answer the question \\  using a single \\ word or phrase} & {\makecell[c]{Condense your answer \\ for each question \\ into a single  word \\ or concise phrase}} & \makecell[l]{Answer with the option's letter from the given choices directly \\ Select the correct answer from the given choices and respond with the letter of the chosen option \\ Determine the correct option from the provided choices and reply with its corresponding letter \\ Pick the correct answer from the listed options and provide the letter of the selected option \\ Identify the correct choice from the options below and respond with the letter of the correct option \\ From the given choices, choose the correct answer and respond with the letter of that choice \\ Choose the right answer from the options and respond with its letter \\ Select the correct answer from the provided options and reply with the letter associated with it \\ From the given choices, select the correct answer and reply with the letter of the chosen option \\ Identify the correct option from the choices provided and respond with the letter of the correct option \\ From the given choices, pick the correct answer and respond by indicating the letter of the correct option} \\ 

\bottomrule
\end{tabular}
}
\end{table*}

\end{document}